\documentclass[letterpaper]{article} % DO NOT CHANGE THIS
\usepackage{aaai24}  % DO NOT CHANGE THIS
\usepackage{times}  % DO NOT CHANGE THIS
\usepackage{helvet}  % DO NOT CHANGE THIS
\usepackage{courier}  % DO NOT CHANGE THIS
\usepackage[hyphens]{url}  % DO NOT CHANGE THIS
\usepackage{graphicx} % DO NOT CHANGE THIS
\usepackage{natbib}  % DO NOT CHANGE THIS AND DO NOT ADD ANY OPTIONS TO IT
\usepackage{caption} % DO NOT CHANGE THIS AND DO NOT ADD ANY OPTIONS TO IT
\usepackage{algorithm}
\usepackage{algorithmic}

\usepackage{newfloat}
\usepackage{listings}
\DeclareCaptionStyle{ruled}{labelfont=normalfont,labelsep=colon,strut=off} % DO NOT CHANGE THIS
\floatstyle{ruled}
\newfloat{listing}{tb}{lst}{}
\floatname{listing}{Listing}
\title{Towards Continual Knowledge Graph Embedding via Incremental Distillation}
\author{
    Jiajun Liu\textsuperscript{\rm 1}\equalcontrib,
    Wenjun Ke\textsuperscript{\rm 1, 2}\equalcontrib \thanks{Corresponding authors.},
    Peng Wang\textsuperscript{\rm 1, 2}\footnotemark[2],
    Ziyu Shang\textsuperscript{\rm 1},\\
    Jinhua Gao\textsuperscript{\rm 3},
    Guozheng Li\textsuperscript{\rm 1},
    Ke Ji\textsuperscript{\rm 1},
    Yanhe Liu\textsuperscript{\rm 1}
}
\affiliations{
    \textsuperscript{\rm 1}School of Computer Science and Engineering, Southeast University\\
    \textsuperscript{\rm 2}Key Laboratory of New Generation Artificial Intelligence Technology and Its Interdisciplinary Applications \\(Southeast University), Ministry of Education, China\\
    \textsuperscript{\rm 3}Institute of Computing Technology, Chinese Academy of Sciences\\
    \{jiajliu, kewenjun, pwang, ziyus1999, liguozheng, keji, liuyanhe\}@seu.edu.cn, gaojinhua@ict.ac.cn
}

\usepackage{bibentry}
\usepackage{amsmath}
\usepackage{multirow}
\usepackage{booktabs}
\usepackage{amssymb}
\usepackage{color}

\begin{document}

\maketitle

\begin{abstract}
Traditional knowledge graph embedding (KGE) methods typically require preserving the entire knowledge graph (KG) with significant training costs when new knowledge emerges. 
To address this issue, the continual knowledge graph embedding (CKGE) task has been proposed to train the KGE model by learning emerging knowledge efficiently while simultaneously preserving decent old knowledge. 
However, the explicit graph structure in KGs, which is critical for the above goal, has been heavily ignored by existing CKGE methods. 
On the one hand, existing methods usually learn new triples in a random order, destroying the inner structure of new KGs. 
On the other hand, old triples are preserved with equal priority, failing to alleviate catastrophic forgetting effectively. 
In this paper, we propose a competitive method for CKGE based on incremental distillation (IncDE), which considers the full use of the explicit graph structure in KGs. 
First, to optimize the learning order, we introduce a hierarchical strategy, ranking new triples for layer-by-layer learning. 
By employing the inter- and intra-hierarchical orders together, new triples are grouped into layers based on the graph structure features. 
Secondly, to preserve the old knowledge effectively, we devise a novel incremental distillation mechanism, which facilitates the seamless transfer of entity representations from the previous layer to the next one, promoting old knowledge preservation. 
Finally, we adopt a two-stage training paradigm to avoid the over-corruption of old knowledge influenced by under-trained new knowledge. 
Experimental results demonstrate the superiority of IncDE over state-of-the-art baselines. 
Notably, the incremental distillation mechanism contributes to improvements of 0.2\%-6.5\% in the mean reciprocal rank (MRR) score. 
More exploratory experiments validate the effectiveness of IncDE in proficiently learning new knowledge while preserving old knowledge across all time steps. 
\end{abstract}

\begin{figure}[t]
\centering
\includegraphics[width=1.00\columnwidth]{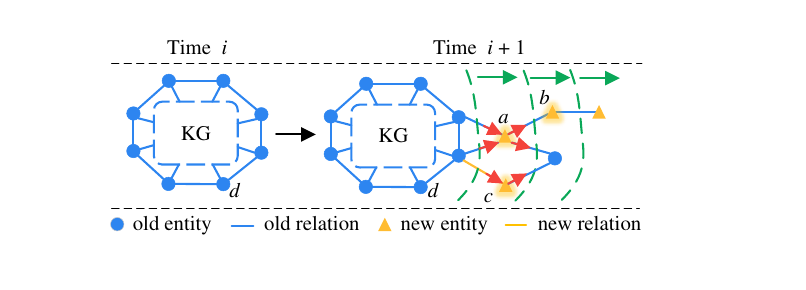}
\caption{Illustration of a growing KG. Two specific learning orders should be considered: entities closer to the old KG should be prioritized ($a$ is prioritised over $b$); entities influenced heavier to new triples (e.g., connecting with more relations) should be prioritized ($a$ is prioritised over $c$).}
\label{figure1}
\end{figure}

\section{Introduction}
Knowledge graph embedding (KGE)~\cite{bordes2013translating, wang2017knowledge, rossi2021knowledge} aims to embed entities and relations from knowledge graphs (KGs) \cite{dong2014knowledge} into continuous vectors in a low-dimensional space, which is crucial for various knowledge-driven tasks, such as question answering~\cite{bordes2014open}, semantic search~\cite{noy2019industry}, and relation extraction~\cite{li2022fastre}. 
Traditional KGE models~\cite{bordes2013translating, trouillon2016complex, sun2019rotate, liu-etal-2020-aprile} only focus on obtaining embeddings of entities and relations in static KGs. 
However, real-world KGs constantly evolve, especially emerging new knowledge, such as new triples, entities, and relations. 
For example, during the evolution of DBpedia~\cite{bizer2009dbpedia} from 2016 to 2018, about 1 million new entities, 2,000 new relations, and 20 million new triples emerged~\cite{dbpedia}. 
Traditionally, when a KG evolves, KGE models need to retrain the models with the entire KG, which is a non-trivial process with huge training costs. 
In domains such as bio-medical and financial fields, it is significant to update the KGE models to support medical assistance and informed market decision-making with rapidly evolving KGs, especially with substantial new knowledge. 

To this end, the continual KGE (CKGE) task has been proposed to alleviate this problem by using only the emerging knowledge for learning~\cite{song2018enriching, daruna2021continual}. 
In comparison with the traditional KGE, the key of CKGE lies in learning emerging knowledge well while preserving old knowledge effectively. 
As shown in Figure~\ref{figure1}, new entities and relations (i.e., the new entity $a$, $b$, and $c$) should be learned to adapt to the new KG. 
Meanwhile, knowledge in the old KG (such as old entity $d$) should be preserved. 
Generally, existing CKGE methods can be categorized into three families: dynamic architecture-based, replay-based, and regularization-based methods. 
Dynamic architecture-based methods~\cite{rusu2016progressive, lomonaco2017core50} preserve all old parameters and learn the emerging knowledge through new architectures. 
However, retaining all old parameters hinders the adaptation of old knowledge to the new knowledge. 
Replay-based methods~\cite{lopez2017gradient, wang2019sentence, kou2020disentangle} replay KG subgraphs to remember old knowledge, but recalling only a portion of the subgraphs leads to the destruction of the overall old graph structure. 
Regularization-based methods~\cite{zenke2017continual, kirkpatrick2017overcoming, cui2023lifelong} aim to preserve old knowledge by adding regularization terms. 
However, only adding regularization terms to the old parameters makes it infeasible to capture new knowledge well.

Despite achieving promising effectiveness, current CKGE methods still perform poorly due to the explicit graph structure of KGs being heavily ignored. 
Meanwhile, previous research has emphasized the crucial role of the graph structure in addressing graph-related continual learning tasks~\cite{zhou2021overcoming,liang2022survey,febrinanto2023graph}. 
Specifically, existing CKGE methods suffer from two main drawbacks: 
(1) First, regarding the new emerging knowledge, current CKGE methods leverage a random-order learning strategy, neglecting the significance of different triples in a KG. 
Previous studies have demonstrated that the learning order of entities and relations can significantly affect continual learning on graphs~\cite{wei2022incregnn}. 
Since knowledge in KGs is organized in a graph structure, a randomized learning order can undermine the inherent semantics conveyed by KGs. 
Hence, it is essential to consider the priority of new entities and relations for effective learning and propagation. 
Figure~\ref{figure1} illustrates an example where entity $a$ should be learned before entity $b$ since the representation of $b$ is propagated through $a$ from the old KG. 
(2) Second, regarding the old knowledge, current CKGE methods treat the memorization at an equal level, leading to inefficient handling of catastrophic forgetting~\cite{kirkpatrick2017overcoming}. 
Existing studies have demonstrated that preserving knowledge by regularization or distillation from important nodes in the topology structure is critical for continuous graph learning~\cite{liu2021overcoming}. 
Therefore, old entities with more essential graph structure features should receive higher preservation priority. 
In Figure~\ref{figure1}, entity $a$ connecting more other entities should be prioritized for preservation at time $i+1$ compared to entity $c$.

In this paper, we propose IncDE, a novel method for the CKGE task that leverages incremental distillation. 
IncDE aims to enhance the capability of learning  emerging knowledge while efficiently preserving old knowledge simultaneously. 
Firstly, we employ hierarchical ordering to determine the optimal learning sequence of new triples. 
This involves dividing the triples into layers and ranking them through the inter-hierarchical and intra-hierarchical orders. 
Subsequently, the ordered emerging knowledge is learned layer by layer. 
Secondly, we introduce a novel incremental distillation mechanism to preserve the old knowledge considering the graph structure effectively. 
This mechanism incorporates the explicit graph structure and employs a layer-by-layer paradigm to distill the entity representation.
Finally, we use a two-stage training strategy to improve the preservation of old knowledge. 
In the first stage, we fix the representation of old entities and relations. 
In the second stage, we train the representation of all entities and relations, protecting the old KG from disruption by under-trained emerging knowledge.

To evaluate the effectiveness of IncDE, we construct three new datasets with varying scales of new KGs. 
Extensive experiments are conducted on both existing and new datasets. 
The results demonstrate that IncDE outperforms all strong baselines. 
Furthermore, ablation experiments reveal that incremental distillation provides a significant performance enhancement. 
Further exploratory experiments verify the ability of IncDE to effectively learn emerging knowledge while efficiently preserving old knowledge.

To sum up, the contributions of this paper are three-fold:
\begin{itemize}

    \item We propose a novel continual knowledge graph embedding framework IncDE, which learns and preserves the knowledge effectively with explicit graph structure. 
    
    \item We propose hierarchical ordering to get an adequate learning order for better learning emerging knowledge. Moreover, we propose incremental distillation and a two stage training strategy to preserve decent old knowledge. 
    
    \item We construct three new datasets based on the scale changes of new knowledge. Experiments demonstrate that IncDE outperforms strong baselines. Notably, incremental distillation improves 0.2\%-6.5\% in MRR. 

\end{itemize}

\begin{figure*}[t]
\centering
\includegraphics[width=1\textwidth]{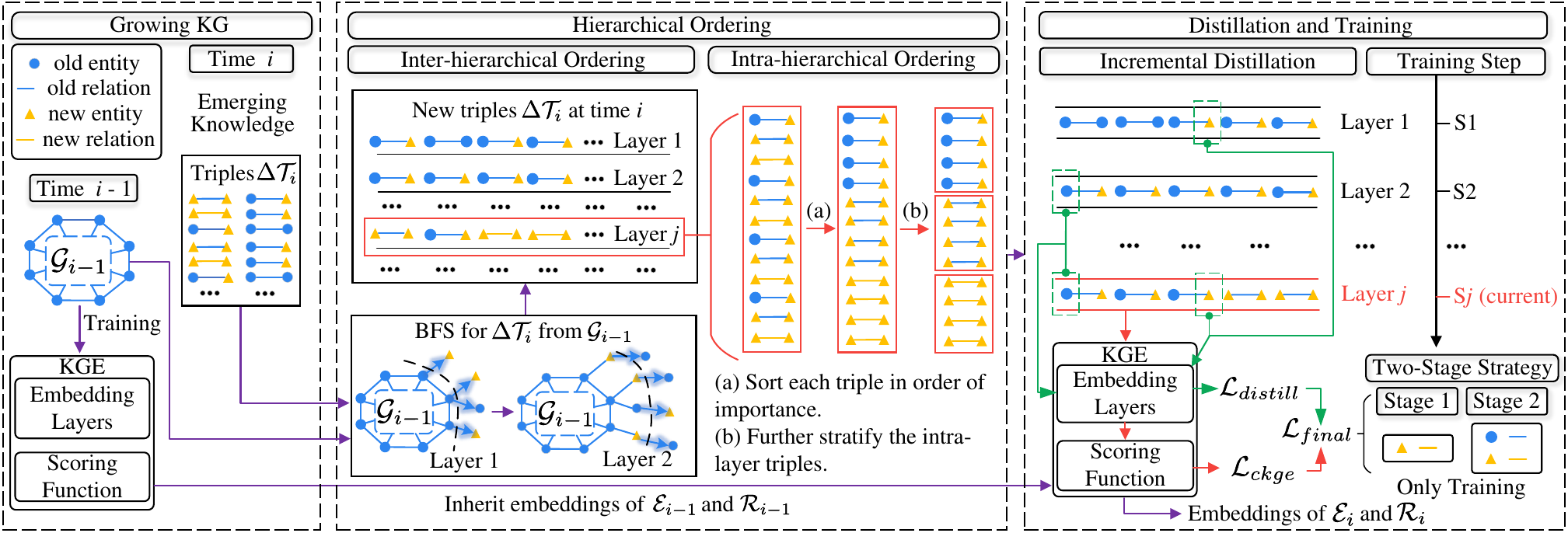} % Reduce the figure size so that it is slightly narrower than the column.
\caption{An overview of our proposed IncDE framework.}
\label{figure2}
\end{figure*}

\section{Related Work}
Different from traditional KGE~\cite{bordes2013translating, trouillon2016complex, kazemi2018simple,pan-wang-2021-hyperbolic-hierarchy,shang2023askrl}, CKGE~\cite{song2018enriching, daruna2021continual} allows KGE models to learn emerging knowledge while remembering the old knowledge. 
Existing CKGE methods can be divided into three categories. 
(1) Dynamic architecture-based methods~\cite{rusu2016progressive, lomonaco2017core50} dynamically adapt to new neural resources to change architectural properties in response to new information and preserve old parameters. 
(2) Memory reply-based methods~\cite{lopez2017gradient, wang2019sentence, kou2020disentangle} retain the learned knowledge by replaying it. 
(3) Regularization-based methods~\cite{zenke2017continual, kirkpatrick2017overcoming, cui2023lifelong} alleviate catastrophic forgetting by imposing constraints on updating neural weights. 
However, these methods overlook the importance of learning new knowledge in an appropriate order for graph data. 
Moreover, they ignore how to preserve appropriate old knowledge for better integration of new and old knowledge. 
Several datasets for CKGE~\cite{hamaguchi2017knowledge, kou2020disentangle, daruna2021continual, cui2023lifelong} have been constructed. 
However, most of them restrict the new triples to contain at least one old entity, neglecting triples without old entities. 
In the evolution of real-world KGs like Wikipedia~\cite{bizer2009dbpedia} and Yago~\cite{10.1145/1242572.1242667}, numerous new triples emerge without any old entities.

\section{Preliminary and Problem Statement}

\subsection{Growing Knowledge Graph}
A knowledge graph (KG) $\mathcal{G} = (\mathcal{E}, \mathcal{R}, \mathcal{T})$ contains the collection of entities $\mathcal{E}$, relations $\mathcal{R}$, and triples $\mathcal{T}$. 
A triple can be denoted as $(h,r,t)\in \mathcal{T}$, where $h$, $r$, and $t$ represent the head entity, the relation, and the tail entity, respectively. 
When a KG grows with emerging knowledge at time $i$, it is denoted as $\mathcal{G}_{i} = (\mathcal{E}_{i}, \mathcal{R}_{i}, \mathcal{T}_{i})$, where $\mathcal{E}_{i}$, $\mathcal{R}_{i}$, $\mathcal{T}_{i}$ are the collection of entities, relations, and triples in $\mathcal{G}_{i}$. 
Moreover, we denote $\Delta \mathcal{T}_{i} = \mathcal{T}_{i} - \mathcal{T}_{i - 1}$, $\Delta \mathcal{E}_{i} = \mathcal{E}_{i} - \mathcal{E}_{i - 1}$ and $\Delta \mathcal{R}_{i} = \mathcal{R}_{i} - \mathcal{R}_{i - 1}$ as new triples, entities, and relations, respectively.

\subsection{Continual Knowledge Graph Embedding}
Given a KG $\mathcal{G}$, knowledge graph embedding (KGE) aims to embed entities and relations into low-dimensional vector space $\mathbb{R}$. 
Given head entity $h \in \mathcal{E}$, relation $r \in \mathcal{R}$, and tail entity $t \in \mathcal{E}$, their embeddings are denoted as $\mathbf{h} \in \mathbb{R}^{d}$, $\mathbf{r} \in \mathbb{R}^{d}$, and $\mathbf{t} \in \mathbb{R}^{d}$, where $d$ is the embedding size. 
A typical KGE model contains embedding layers and a scoring function. 
Embedding layers generate vector representations for entities and relations, while the scoring function assigns scores to each triple in the training stage.

Given a growing KG $\mathcal{G}_{i}$ at time $i$, continual knowledge graph embedding (CKGE) aims to update the embeddings of old entities $\mathcal{E}_{i-1}$ and relations $\mathcal{R}_{i-1}$ while obtaining the embeddings of new entities $\Delta \mathcal{E}_{i}$ and relations $\Delta \mathcal{R}_{i}$. 
Finally, embeddings of all entities $\mathcal{E}_{i}$ and relations $\mathcal{R}_{i}$ are obtained.

\section{Methodology}
\subsection{Framework Overview}
The framework of IncDE is depicted in Figure~\ref{figure2}. 
Initially, when emerging knowledge appears at time $i$, IncDE performs hierarchical ordering on new triples $\Delta \mathcal{T}_{i}$. 
Specifically, inter-hierarchical ordering is employed to divide $\Delta \mathcal{T}_{i}$ into multiple layers using breadth-first search (BFS) expansion from the old graph $\mathcal{G}_{i-1}$. 
Subsequently, intra-hierarchical ordering is applied within each layer to further sort and divide the triples. 
Then, the grouped $\Delta \mathcal{T}_{i}$ is trained layer by layer, with the embeddings of $\mathcal{E}_{i-1}$ and $\mathcal{R}_{i-1}$ inherited from the KGE model in previous time $i-1$. 
During training, incremental distillation is introduced. 
Precisely, if an entity in  layer $j$ has appeared in a previous layer, its representation is distilled with the closest layer to the current one. 
Additionally, a two-stage training strategy is proposed. 
In the first stage, only the representations of new entities $\Delta \mathcal{E}_{i}$ and relations $\Delta \mathcal{R}_{i}$ are trained. 
In the second stage, all entities $\mathcal{E}_{i}$ and relations $\mathcal{R}_{i}$ are trained in the training process. 
Finally, the embeddings of $\mathcal{E}_{i}$ and $\mathcal{R}_{i}$ at time $i$ are obtained.

\subsection{Hierarchical Ordering}
To enhance the learning of the graph structure for emerging knowledge, we first order the triples $\Delta \mathcal{T}_{i}$ at time $i$ in an inter-hierarchical way and an intra-hierarchical way, based on the importance of entities and relations, as shown in Figure~\ref{figure2}. 
Ordering processes can be pre-calculated to reduce training time. 
Then, we learn the new triples $\Delta \mathcal{T}_{i}$ layer by layer and in order. 
The specific ordering strategies are as follows.

\subsubsection{Inter-Hierarchical Ordering}
For inter-hierarchical ordering, we split all new triples $\Delta \mathcal{T}_{i}$ into multiple layers $l_{1}, l_{2}, ..., l_{n}$ at time $i$. 
Since the representations of new entities $\Delta \mathcal{E}_{i}$ are propagated from the representations of the old entities $\mathcal{E}_{i-1}$ and old relations $\mathcal{R}_{i-1}$, we split new triples $\Delta \mathcal{T}_{i}$ based on the distance between new entities $\Delta \mathcal{E}_{i}$ and  old graph $\mathcal{G}_{i-1}$. 
We use the bread-first search (BFS) algorithm to progressively partition $\Delta \mathcal{T}_{i}$ from $\mathcal{G}_{i-1}$. 
First, we take the old graph as $l_{0}$. 
Then, we take all the new triples that contain old entities as the next layer, $l_{1}$. 
Next, we treat the new entities in $l_{1}$ as the seen old entities. 
Repeat the above two processes until no triples can be added to a new layer. 
Finally, we use all remaining triples as the final layer. 
This way, we initially divide all the new triples $\Delta \mathcal{T}_{i}$ into multiple layers. 

\subsubsection{Intra-Hierarchical Ordering}
The importance of the triples in graph structure is also critical to the order in which entities $\mathcal{E}_{i}$ and relations $\mathcal{R}_{i}$ are learned or updated at time $i$. 
So for the triples of each layer, we further order them based on the importance of entities and relations in the graph structure, as shown in Figure~\ref{figure2} (a). 
To measure the importance of entities $\mathcal{E}_{i}$ in the new triples $\Delta \mathcal{T}_{i}$, we first calculate the node centrality of an entity $e \in \mathcal{E}_{i}$ as $f_{nc}(e)$ as follow: 
\begin{equation}
    f_{nc}(e) = \frac{f_{neighbor}(e)}{N - 1}
    \label{e3}
\end{equation}
where $f_{neighbor}(e)$ denotes the number of the neighbors of $e$, and $N$ denotes the number of entities in the new triples $\Delta \mathcal{T}_{i}$ at time $i$. 
Then, in order to measure the importance of relations $\mathcal{R}_{i}$ in the triples of each layer, we compute the betweenness centrality of a relation $r \in \mathcal{R}_{i}$ as $f_{bc}(r)$: 
\begin{equation}
    f_{bc}(r) =\sum_{s,t \in \mathcal{E}_{i}, s \neq t} \frac{\sigma(s, t|r)}{\sigma(s, t)}
    \label{e4}
\end{equation}
where $\sigma(s, t)$ is the number of shortest paths between $s$ and $t$ in the new triples $\Delta \mathcal{T}_{i}$, and $\sigma(s, t|r)$ is the number of $\sigma(s, t)$ passing through relation $r$. 
Specifically, we only compute $f_{nc}$ and $f_{bc}$ of emerging KGs, avoiding the graph being excessive. 
To obtain the importance of the triple $(h, r, t)$ in each layer, we compute the node centrality of the head entity $h$, the node centrality of the tail entity $t$, and the betweenness centrality of the relation $r$ in this triple. 
Considering the overall significance of entities and relations within the graph structure, we adopt $f_{nc}$ and $f_{bc}$ together. 
The final importance of each triple can be calculated as: 
\begin{equation}
    IT_{(h, r, t)} = max(f_{nc}(h), f_{nc}(t)) + f_{bc}(r)
    \label{e5}
\end{equation}
We sort the triples of each layer according to the values of their $IT$ values. 
The utilization of intra-hierarchical ordering guarantees the prioritization of triples that are important to the graph structure in each layer. 
This, in turn, enables more effective learning of the structure of the new graph. 

Moreover, the intra-hierarchical ordering can help further split the intra-layer triples, as shown in Figure~\ref{figure2} (b). 
Since the number of triples in each layer is determined by the size of the new graph, it could be too large to learn. 
To prevent the number of triples in a particular layer from being too large, we set the maximum number of triples in each layer to be $M$. 
If the number of triples in one layer exceeds $M$, it can split into several layers not exceeding $M$ triples in the intra-hierarchical ordering.

\begin{table*}[htb!]
\centering
\setlength{\tabcolsep}{0.9mm}
% \small
% \Large
\begin{tabular}{lcccccccccccccccc}
\hline
\multirow{3}{*}{Dataset} & \multicolumn{3}{c}{Time 1} & \multicolumn{3}{c}{Time 2} & \multicolumn{3}{c}{Time 3} & \multicolumn{3}{c}{Time 4} & \multicolumn{3}{c}{Time 5} \\
 & $N_{E}$ & $N_{R}$ & $N_{T}$ & $N_{E}$ & $N_{R}$ & $N_{T}$ & $N_{E}$ & $N_{R}$ & $N_{T}$ & $N_{E}$ & $N_{R}$ & $N_{T}$ & $N_{E}$ & $N_{R}$ & $N_{T}$ \\
\hline
ENTITY & 2,909 & 233 & 46,388 & 5,817 & 236 & 72,111 & 8,275 & 236 & 73,785 & 11633 & 237 & 70,506 & 14,541 & 237 & 47,326 \\
RELATION & 11,560 & 48 & 98,819 & 13,343 & 96 & 93,535 & 13,754 & 143 & 66,136 & 14,387 & 190 & 30,032 & 14,541 & 237 & 21,594 \\
FACT & 10,513 & 237 & 62,024 & 12,779 & 237 & 62,023 & 13,586 & 237 & 62,023 & 13,894 & 237 & 62,023 & 14,541 & 237 & 62,023 \\
HYBRID & 8,628 & 86 & 57,561 & 10,040 & 102 & 20,873 & 12,779 & 151 & 88,017 & 14,393 & 209 & 103,339 & 14,541 & 237 & 40,326 \\
GraphEqual & 2,908 & 226 & 57,636 & 5,816 & 235 & 62,023 & 8,724 & 237 & 62,023 & 11,632 & 237 & 62,023 & 14,541 & 237 & 66,411\\
GraphHigher & 900 & 197 & 10,000 & 1,838 & 221 & 20,000 & 3,714 & 234 & 40,000 & 7,467 & 237 & 80,000 & 14,541 & 237 & 160,116 \\
GraphLower & 7,505 & 237 & 160,000 & 11,258 & 237 & 80,000 & 13,134 & 237 & 40,000 & 14,072 & 237 & 20,000 & 14,541 & 237 & 10,116 \\
\hline
\end{tabular}
\caption{The statistics of datasets. $N_{E}$, $N_{R}$ and $N_{T}$ denote the number of cumulative entities, cumulative relations and current triples at each time $i$.}
\label{t1}
\end{table*}

\subsection{Distillation and Training}
After hierarchical ordering, we train new triples $\Delta \mathcal{T}_{i}$ layer by layer at time $i$. 
We take TransE~\cite{bordes2013translating} as the base KGE model. 
When training the $j$-th layer ($j > 0$), the loss for the original TransE model is: 
\begin{equation}
    \mathcal{L}_{ckge} = \sum_{(h, r, t) \in l_{j}} max(0, f(h, r, t) - f(h', r, t') + \gamma)
    \label{e2}
\end{equation}
where $(h', r, t')$ is the negative triple of $(h, r, t) \in l_{j}$, and $f(h, r, t) = |h + r - t|_{L1/L2}$ is the score function of TransE. 
We inherit the embeddings of old entities $\mathcal{E}_{i - 1}$ and relations $\mathcal{R}_{i - 1}$ from the KGE model at time $i-1$ and randomly initialize the embeddings of new entities $\Delta \mathcal{E}_{i}$ and relations $\Delta \mathcal{R}_{i}$. 
During training, we use incremental distillation to preserve the old knowledge. 
Further, we propose a two-stage training strategy to prevent the embeddings of old entities and relations from being overly corrupted at the start of training.

\subsubsection{Incremental Distillation}
In order to alleviate catastrophic forgetting of the entities learned in previous layers, inspired by the knowledge distillation for KGE models~\cite{wang2021mulde, zhu2022dualde, liu2023iterde}, we distill the entity representation in the current layer with the entities that have appeared in previous layers as shown in Figure~\ref{figure2}. 
% Compared to previous KGE distillation methods that distill final distribution, we distill the intermediate hidden states to preserve essential features of old knowledge. 
Specifically, if entity $e$ in the $j$-th ($j > 0$) layer has appeared in a previous layer, we distill it with the representation of $e$ from the nearest layer. 
The loss of distillation for entity $e_{k}$ ($k \in [1, |\mathcal{E}_{i}|]$) is: 
\begin{equation}
\mathcal{L}_{distill}^{k} = \left\{\begin{matrix} 
  \frac{1}{2}(\mathbf{e'}_{k} - \mathbf{e}_{k})^2, &|\mathbf{e'}_{k} - \mathbf{e}_{k}| \le 1 \\  
  |\mathbf{e'}_{k} - \mathbf{e}_{k}|-\frac{1}{2},&|\mathbf{e'}_{k} - \mathbf{e}_{k}| > 1
\end{matrix}\right. 
\label{e6}
\end{equation}
where $\mathbf{e}_{k}$ denotes the representation of entity $e_{k}$ in layer $j$, $\mathbf{e'}_{k}$ denotes the representation of entity $e_{k}$ from the nearest previous layer. 
By distilling entities that have appeared in previous layers, we remember old knowledge efficiently. 
However, different entities should have different levels of memory for past representations. 
Entities with higher importance in the graph structure should be prioritized and preserved to a greater extent during distillation. 
Besides the node centrality of the entity $f_{nc}$, similar to the betweenness centrality of the relation, we define the betweenness centrality $f_{bc}(e)$ of an entity $e$ at time $i$ as: 
\begin{equation}
    f_{bc}(e) =\sum_{s,t \in \mathcal{E}_{i}, s \neq t} \frac{\sigma(s, t|e)}{\sigma(s, t)}
    \label{e7}
\end{equation}
We combine $f_{bc}(e)$ and $f_{nc}(e)$ to evaluate the importance of an entity $e$. 
Concretely, when training the $j$-th layer, for each new entity $e_{k}$ appearing at the time $i$, we compute $f_{bc}(e_{k})$ and $f_{nc}(e_{k})$ to get the preliminary weight $\lambda_{k}$ as: 
\begin{equation}
    \lambda_{k} = \lambda_{0} \cdot (f_{bc}(e_{k}) + f_{nc}(e_{k}))
    \label{e8}
\end{equation}
where $\lambda_{0}$ is $1$ for new entities that have already appeared in previous layers, and $\lambda_{0}$ is 0 for new entities that have not appeared. 
At the same time, we learn a matrix $\mathbf{W} \in \mathbb{R}^{1 \times |\mathcal{E}_{i}|}$ to dynamically change the weights of distillation loss for different entities. 
The dynamic distillation weights is: 
\begin{equation}
    [\lambda_{1}^{'}, \lambda_{2}^{'}, ..., \lambda_{|\mathcal{E}_{i}|}^{'}] = [\lambda_{1}, \lambda_{2}, ..., \lambda_{|\mathcal{E}_{i}|}] \circ \mathbf{W}
    \label{e9}
\end{equation}
where $\circ$ denotes the Hadamard product. The final distillation loss for each layer $j$ at the time $i$ is:
\begin{equation}
    \mathcal{L}_{distill} = \sum_{k=1}^{|\mathcal{E}_{i}|} \lambda_{k}^{'} \cdot \mathcal{L}_{distill}^{k}
    \label{e10}
\end{equation}
When training the $j$-th layer, the final loss function can be calculated as: 
\begin{equation}
    \mathcal{L}_{final} = \mathcal{L}_{ckge} + \mathcal{L}_{distill}
    \label{e11}
\end{equation}
After layer-by-layer training for new triples $\Delta \mathcal{T}_{i}$, all representations of entities $\mathcal{E}_{i}$ and relations $\mathcal{R}_{i}$ are obtained. 

\subsubsection{Two-Stage Training}
During the training process, when incorporating the new triples $\Delta \mathcal{T}_{i}$ into the existing graph $\mathcal{G}_{i-1}$ at time $i$, the embeddings of old entities and relations that are not present in the new triples $\Delta \mathcal{T}_{i}$ remain unchanged. 
However, the embeddings of old entities and relations that are included in the new triples $\Delta \mathcal{T}_{i}$ are updated. 
Therefore, in the initial stage of each time $i$, part of the representations of entities $\mathcal{E}_{i - 1}$ and relations $\mathcal{R}_{i - 1}$ in the old graph $\mathcal{G}_{i - 1}$ will be corrupted by the new entities $\Delta \mathcal{E}_{i}$ and relations $\Delta \mathcal{R}_{i}$ that are not fully trained. 
To solve this problem, IncDE uses a two-stage training strategy to preserve the knowledge in the old graph better, as shown in Figure~\ref{figure2}. 
In the first training stage, IncDE freezes the embeddings of all old entities $\mathcal{E}_{i - 1}$ and relations $\mathcal{R}_{i - 1}$ and trains only the embeddings of new entities $\Delta \mathcal{E}_{i}$ and relations $\Delta \mathcal{R}_{i}$. 
Then, IncDE trains the embeddings of all entities $\mathcal{E}_{i}$ and relations $\mathcal{R}_{i}$ in the new graph in the second training stage. 
With the two-stage training strategy, IncDE prevents the structure of the old graph from disruption by new triples in the early training phase. 
At the same time, the representations of entities and relations in the old graph and those in the new graph can be better adapted to each other during training.

\section{Experiments}
\subsection{Experimental Setup}
\subsubsection{Datasets}
We use seven datasets for CKGE, including four public datasets~\cite{cui2023lifelong}: ENTITY, RELATION, FACT, HYBRID, as well as three new datasets constructed by us: GraphEqual, GraphHigher, and GraphLower. 
In ENTITY, RELATION, and FACT, the number of entities, relations, and triples increases uniformly at each time step. 
In HYBRID, the sum of entities, relations, and triples increases uniformly over time. 
However, these datasets constrain knowledge growth, requiring new triples to include at least one existing entity. 
To address this limitation, we relax these constraints and construct three new datasets: GraphEqual, GraphHigher, and GraphLower. 
In GraphEqual, the number of triples consistently increases by the same increment at each time step. 
In GraphHigher and GraphLower, the increments of triples become higher and lower, respectively. 
Detailed statistics for all datasets are presented in Table~\ref{t1}. 
The time step is set to 5. 
The train, valid, and test sets are allocated 3:1:1 for each time step. 
The datasets are available at https://github.com/seukgcode/IncDE.

\subsubsection{Baselines}
We select two kinds of baseline models: non-continual learning methods and continual learning-based methods. 
First, we select a non-continual learning method, Fine-tune~\cite{cui2023lifelong}, which is fine-tuned with the new triples each time. 
Then, we select three kinds of continual learning-based methods: dynamic architecture-based, memory replay-based baselines, and regularization-based. 
Specifically, the dynamic architecture-based methods are PNN~\cite{rusu2016progressive} and CWR~\cite{lomonaco2017core50}. 
The memory replay-based methods are GEM~\cite{lopez2017gradient}, EMR~\cite{wang2019sentence}, and DiCGRL~\cite{kou2020disentangle}. 
The regularization-based methods are SI~\cite{zenke2017continual}, EWC~\cite{kirkpatrick2017overcoming}, and LKGE~\cite{cui2023lifelong}.

\subsubsection{Metrics}
We evaluate our model performance on the link prediction task. 
Particularly, we replace the head or tail entity of the triples in the test set with all other entities and then compute and rank the scores for each triple. 
Then, we compute MRR, Hits@1, and Hits@10 as metrics. The higher the MRR, Hits@1, Hits@3, and Hits@10, the better the model works. 
At time $i$, we use the mean of the metrics tested on all test sets at the time [1, $i$] as the final metric. 
The main results are obtained from the model generated at the last time.

\begin{table*}[htb!]
\centering
\setlength{\tabcolsep}{1mm}
% \small
% \Large
\begin{tabular}{l|ccc|ccc|ccc|ccc|ccc}
\hline
\multirow{3}{*}{Method} & \multicolumn{3}{c|}{ENTITY} & \multicolumn{3}{c|}{RELATION} & \multicolumn{3}{c|}{FACT} & \multicolumn{3}{c|}{HYBRID} & \multicolumn{3}{c}{GraphEqual} \\
 & MRR & H@1 & H@10 & MRR & H@1 & H@10 & MRR & H@1 & H@10 & MRR & H@1 & H@10 & MRR & H@1 & H@10 \\
\hline
Fine-tune & 0.165 & 0.085 & 0.321  & 0.093 & 0.039 & 0.195 & 0.172 & 0.090 & 0.339 & 0.135 & 0.069 & 0.262 & 0.183 & 0.096 & 0.358 \\
\hline
PNN & 0.229 & 0.130 & \underline{0.425} & 0.167 & 0.096 & 0.305 & 0.157 & 0.084 & 0.290 & 0.185 & 0.101 & 0.349 & 0.212 & \underline{0.118} & 0.405 \\
CWR & 0.088 & 0.028 & 0.202 & 0.021 & 0.010 & 0.043 & 0.083 & 0.030 & 0.192 & 0.037 & 0.015 & 0.077 & 0.122 & 0.041 & 0.277 \\
\hline
GEM & 0.165 & 0.085 & 0.321 & 0.093 & 0.040 & 0.196 & 0.175 & 0.092 & 0.345 & 0.136 & 0.070 & 0.263 & 0.189 & 0.099 & 0.372  \\
EMR & 0.171 & 0.090 & 0.330 & 0.111 & 0.052 & 0.225 & 0.171 & 0.090 & 0.337 & 0.141 & 0.073 & 0.267 & 0.185 & 0.099 & 0.359 \\
DiCGRL & 0.107 & 0.057 & 0.211 & 0.133 & 0.079 & 0.241 & 0.162 & 0.084 & 0.320 & 0.149 & 0.083 & 0.277 & 0.104 & 0.040 & 0.226 \\
\hline
SI & 0.154 & 0.072 & 0.311 & 0.113 & 0.055 & 0.224 & 0.172 & 0.088 & 0.343 & 0.111 & 0.049 & 0.229 & 0.179 & 0.092 & 0.353 \\
EWC & 0.229 & 0.130 & 0.423 & 0.165 & 0.093 & 0.306 & 0.201 & 0.113 & 0.382 & 0.186 & 0.102 & 0.350 & 0.207 & 0.113 & 0.400 \\
LKGE & \underline{0.234} & \underline{0.136} & \underline{0.425} & \underline{0.192} & \underline{0.106} & \underline{0.366} & \underline{0.210} & \underline{0.122} & \underline{0.387} & \underline{0.207} & \underline{0.121} & \underline{0.379} & \underline{0.214} & \underline{0.118} & \underline{0.407} \\
\hline
\textbf{IncDE} & \textbf{0.253} & \textbf{0.151} & \textbf{0.448} & \textbf{0.199} & \textbf{0.111} & \textbf{0.370} & \textbf{0.216} & \textbf{0.128} & \textbf{0.391} & \textbf{0.224} & \textbf{0.131} & \textbf{0.401} & \textbf{0.234} & \textbf{0.134} & \textbf{0.432}  \\
\hline
\end{tabular}
\caption{Main experimental results on ENTITY, RELATION, FACT, HYBRID, and GraphEqual. The bold scores indicate the best results and underlined scores indicate the second best results.}
\label{t2}
\end{table*}

\begin{table}[htb!]
\centering
\setlength{\tabcolsep}{1mm}
% \small
% \Large
\begin{tabular}{l|ccc|ccc}
\hline
\multirow{3}{*}{Method} & \multicolumn{3}{c|}{GraphHigher} & \multicolumn{3}{c}{GraphLower} \\
 & MRR & H@1 & H@10 & MRR & H@1 & H@10 \\
\hline
Fine-tune & 0.198 & 0.108 & 0.375 & 0.185 & 0.098 & 0.363 \\
\hline
PNN & 0.186 & 0.097 & 0.364 & \underline{0.213} & \underline{0.119} & \underline{0.407} \\
CWR & 0.189 & 0.096 & 0.374 & 0.032 & 0.005 & 0.080 \\
\hline
GEM & 0.197 & 0.109 & 0.372 & 0.170 & 0.084 & 0.346 \\
EMR & 0.202 & 0.113 & 0.379 & 0.188 & 0.101 & 0.362 \\
DiCGRL & 0.116 & 0.041 & 0.242 & 0.102 & 0.039 & 0.222 \\
\hline
SI & 0.190 & 0.099 & 0.371 & 0.186 & 0.099 & 0.366 \\
EWC & 0.198 & 0.106 & \underline{0.385} & 0.210 & 0.116 & 0.405 \\
LKGE & \underline{0.207} & \underline{0.120} & 0.382 & 0.210 & 0.116 & 0.403 \\
\hline
\textbf{IncDE} & \textbf{0.227} & \textbf{0.132} & \textbf{0.412} & \textbf{0.228} & \textbf{0.129} & \textbf{0.426} \\
\hline
\end{tabular}
\caption{Main experimental results on GraphHigher and GraphLower.}
\label{t3}
\end{table}

\subsubsection{Settings}
All experiments are implemented on the NVIDIA RTX 3090Ti GPU with the PyTorch~\cite{NEURIPS2019_9015}. 
In all experiments, we set TransE~\cite{bordes2013translating} as the base KGE model and the max size of time $i$ as 5. 
The embedding size for entities and relations is 200. 
We tune the batch size in [512, 1024, 2048]. 
We choose Adam as the optimizer and set the learning rate from [1e-5, 1e-4, 1e-3]. 
In our experiments, we set the max number of triples in each layer $M$ in [512, 1024, 2048]. 
To ensure fairness, all experimental results are averages of 5 running times.

\begin{table*}[htb!]
\centering
\setlength{\tabcolsep}{1.4mm}
\small
% \Large
\begin{tabular}{l|ccc|ccc|ccc|ccc|ccc}
\hline
\multirow{3}{*}{Method} & \multicolumn{3}{c|}{ENTITY} & \multicolumn{3}{c|}{RELATION} & \multicolumn{3}{c|}{FACT} & \multicolumn{3}{c|}{HYBRID} & \multicolumn{3}{c}{GraphEqual} \\
 & MRR & H@1 & H@10 & MRR & H@1 & H@10 & MRR & H@1 & H@10 & MRR & H@1 & H@10 & MRR & H@1 & H@10 \\
\hline
IncDE w/o HO & 0.248 & 0.148 & 0.441 & 0.186 & 0.105 & 0.344 & 0.197 & 0.119 & 0.347 & 0.210 & 0.122 & 0.380 & 0.230 & 0.131 & 0.426 \\
IncDE w/o ID & 0.188 & 0.099 & 0.354 & 0.134 & 0.070 & 0.254 & 0.167 & 0.090 & 0.321 & 0.185 & 0.105 & 0.340 & 0.199 & 0.107 & 0.383 \\
IncDE w/o TS & 0.250 & 0.149 & 0.444 & 0.186 & 0.099 & 0.354 & 0.213 & 0.126 & 0.389 & 0.220 & 0.127 & 0.397 & 0.231 & 0.132 & 0.430 \\
\textbf{IncDE} & \textbf{0.253} & \textbf{0.151} & \textbf{0.448} & \textbf{0.199} & \textbf{0.111} & \textbf{0.370} & \textbf{0.216} & \textbf{0.128} & \textbf{0.391} & \textbf{0.224} & \textbf{0.131} & \textbf{0.401} & \textbf{0.234} & \textbf{0.134} & \textbf{0.432} \\
\hline
\end{tabular}
\caption{Ablation experimental results on ENTITY, RELATION, FACT, HYBRID and GraphEqual. HO is the hierarchical ordering. ID is the incremental distillation. TS is the two-stage. We learn the new KG in randomized order w/o HO.}
\label{t4}
\end{table*}

\begin{table}[htb!]
\setlength\tabcolsep{1.4mm}
\centering
\small
% \Large
\begin{tabular}{l|ccc|ccc}
\hline
\multirow{3}{*}{Method} & \multicolumn{3}{c|}{GraphHigher} & \multicolumn{3}{c}{GraphLower} \\
 & MRR & H@1 & H@10 & MRR & H@1 & H@10 \\
\hline
IncDE w/o HO & 0.221 & 0.129 & 0.405 & 0.224 & 0.126 & 0.424 \\
IncDE w/o ID & 0.225 & 0.131 & 0.410 & 0.196 & 0.105 & 0.377 \\
IncDE w/o TS & 0.225 & 0.130 & 0.408 & 0.225 & 0.128 & 0.423 \\
\textbf{IncDE} & \textbf{0.227} & \textbf{0.132} & \textbf{0.412} & \textbf{0.228} & \textbf{0.129} & \textbf{0.426} \\
\hline
\end{tabular}
\caption{Ablation experimental results on GraphHigher and GraphLower.}
\label{t5}
\end{table}

\subsection{Results}
\subsubsection{Main Results}
The results of the main experiments on the seven datasets are reported in Table~\ref{t2} and Table~\ref{t3}. 

Firstly, it is worth noting that IncDE exhibits a considerable improvement when compared to Fine-tune. 
Specifically, IncDE demonstrates enhancements ranging from 2.9\%-10.6\% in MRR, 2.4\%-7.2\% in Hits@1, and 3.7\%-17.5\% in Hits@10 compared to Fine-tune. 
The results suggest that direct fine-tuning leads to catastrophic forgetting. 

Secondly, IncDE outperforms all CKGE baselines. 
Notably, IncDE achieves improvements of 1.5\%-19.6\%, 1.0\%-12.4\%, and 1.9\%-34.6\%, respectively, in MRR, Hits@1, and Hits@10 compared to dynamic architecture-based approaches (PNN and CWR). 
Compared to replay-based baselines (GEM, EMR, and DiCGRL), IncDE improves 2.5\%-14.6\%, 1.9\%-9.4\%, and 3.3\%-23.7\% in MRR, Hits@1, and Hits@10. 
Moreover, IncDE obtains 0.6\%-11.3\%, 0.5\%-8.2\%, and 0.4\%-17.2\% improvements in MRR, Hits@1, and Hits@10 compared to regularization-based methods (SI, EWC, and LKGE). 
These results demonstrate the superior performance of IncDE on growing KGs.

Thirdly, IncDE exhibits distinct improvements across different types of datasets when compared to the strong baselines. 
In datasets with equal growth of knowledge (ENTITY, FACT, RELATION, HYBRID, and GraphEqual), IncDE has an average improvement of 1.4\% in MRR over the state-of-the-art methods. 
In datasets with unequal growth of knowledge (GraphHigher and GraphLower), IncDE demonstrates an improvement of 1.8\%-2.0\% in MRR over the optimal methods. 
It means that IncDE is particularly well-suited for scenarios involving unequal knowledge growth. 
Notably, when dealing with a more real-scenario-aware dataset, GraphHigher, where a substantial amount of new knowledge emerges, IncDE demonstrates the most apparent advantages compared to other strongest baselines by 2.0\% in MRR. 
It indicates that IncDE performs well when a substantial amount of new knowledge is emerging. 
Therefore, we verify the scalability of IncDE in datasets (GraphHigher, GraphLower, and GraphEqual) with varying sizes (triples from 10K to 160K, from 160K to 10K, and the remaining 62K). 
In particular, we observe that IncDE only improves by 0.6\%-0.7\% in MRR on RELATION and FACT compared to the best results among all baselines, where the improvements are insignificant as other datasets. 
This can be attributed to the limited growth of new entities in these two datasets, indicating that IncDE is highly adaptable to situations where the number of entities varies significantly. 
In real life, the number of relations between entities remains relatively stable, while it is the entities themselves that appear in large numbers. 
This is where IncDE excels in its adaptability. 
With its robust capabilities, IncDE can effectively handle the multitude of entities and their corresponding relations, ensuring seamless integration and efficient processing.

\begin{figure}[t]
\includegraphics[width=0.46\textwidth]{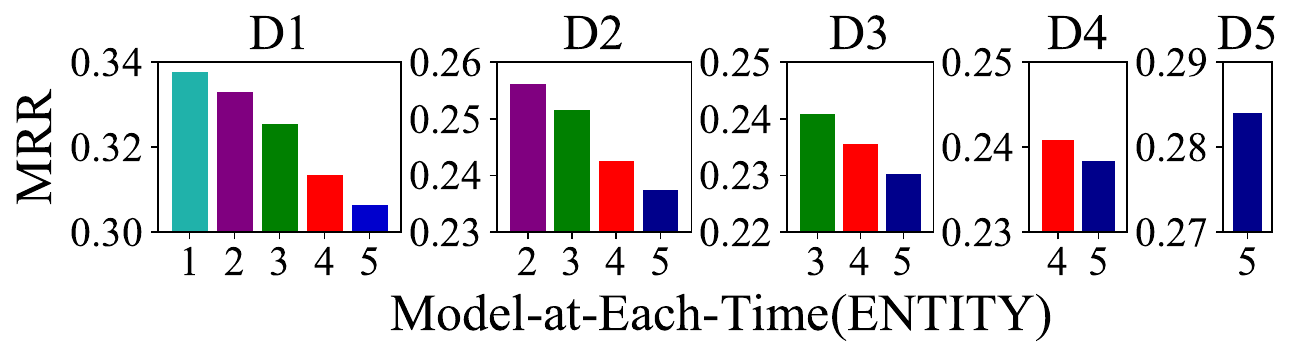} \label{Fig.3(a)} \\
\includegraphics[width=0.46\textwidth]{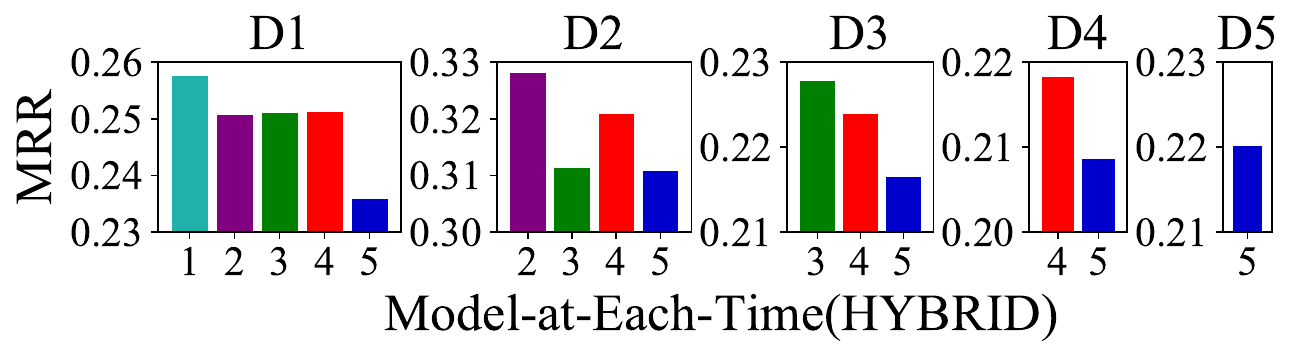} \label{Fig.3(b)} \\
\includegraphics[width=0.46\textwidth]{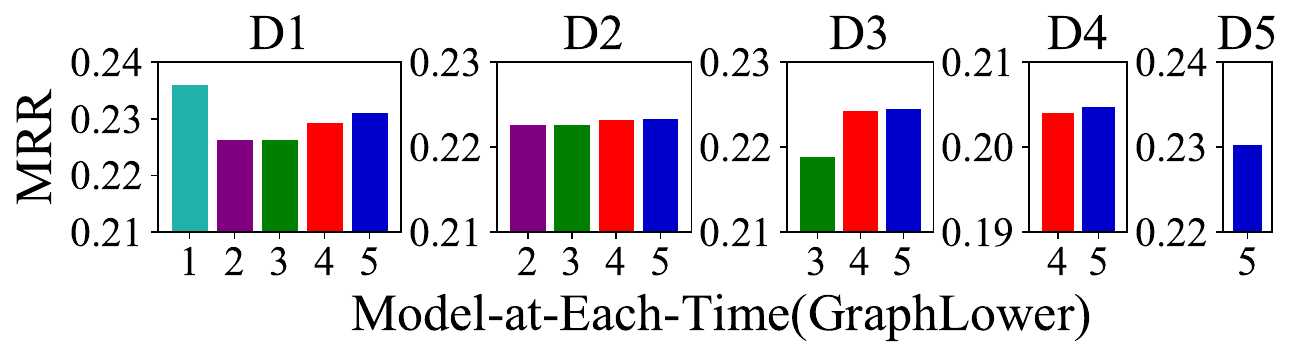} \label{Fig.3(c)}
\caption{Effectiveness of IncDE at Each Time on ENTITY, HYBRID, and GraphLower. Different colors represent the performance of models generated at different times. D$i$ denotes the test set at time $i$.}
\label{figure3}
\end{figure}

\subsubsection{Ablation Experiments}
We investigate the effects of hierarchical ordering, incremental distillation, and the two-stage training strategy, as depicted in Table~\ref{t4} and Table~\ref{t5}. 
Firstly, when we remove the incremental distillation, there is a significant decrease in the model performance. 
Specifically, the metrics decrease by 0.2\%-6.5\% in MRR,  0.1\%-5.2\% in Hits@1, and 0.2\%-11.6\% in Hits@10. 
These findings highlight the crucial role of incremental distillation in effectively preserving the structure of the old graph while simultaneously learning the representation of the new graph. 
Secondly, there is a slight decline in model performance when we eliminate the hierarchical ordering and two-stage training strategy. 
Specifically, the metrics of MRR decreased by 0.2\%-1.8\%, Hits@1 decreased by 0.1\%-1.8\%, and Hits@10 decreased by 0.2\%-4.4\%. 
The results show that the hierarchical ordering and the two-stage training improve the performance of IncDE.

\subsubsection{Performance of IncDE in Each Time}
Figure \ref{figure3} shows how well IncDE remembers old knowledge at different times. 
First, we observe that on several test data (D1, D2, D3, D4 in ENTITY; D3, D4 in HYBRID), the performance of IncDE decreases slightly by 0.2\%-3.1\% with increasing time. 
In particular, the performance of IncDE does not undergo significant degradation on several datasets, such as D1 of HYBRID (Time 2 to Time 4) and D2 of GraphLower (Time 2 to Time 5). 
It means that IncDE can remember old knowledge well on most datasets. 
Second, on a few datasets, the performance of IncDE unexpectedly gains as it continues to be trained. 
Specifically, the performance of IncDE gradually increases by 0.6\% on D3 of GraphLower in MRR. 
This demonstrates that IncDE learns emerging knowledge well and enhances the old knowledge with emerging knowledge.

\begin{figure}[t]
\includegraphics[width=0.46\textwidth]{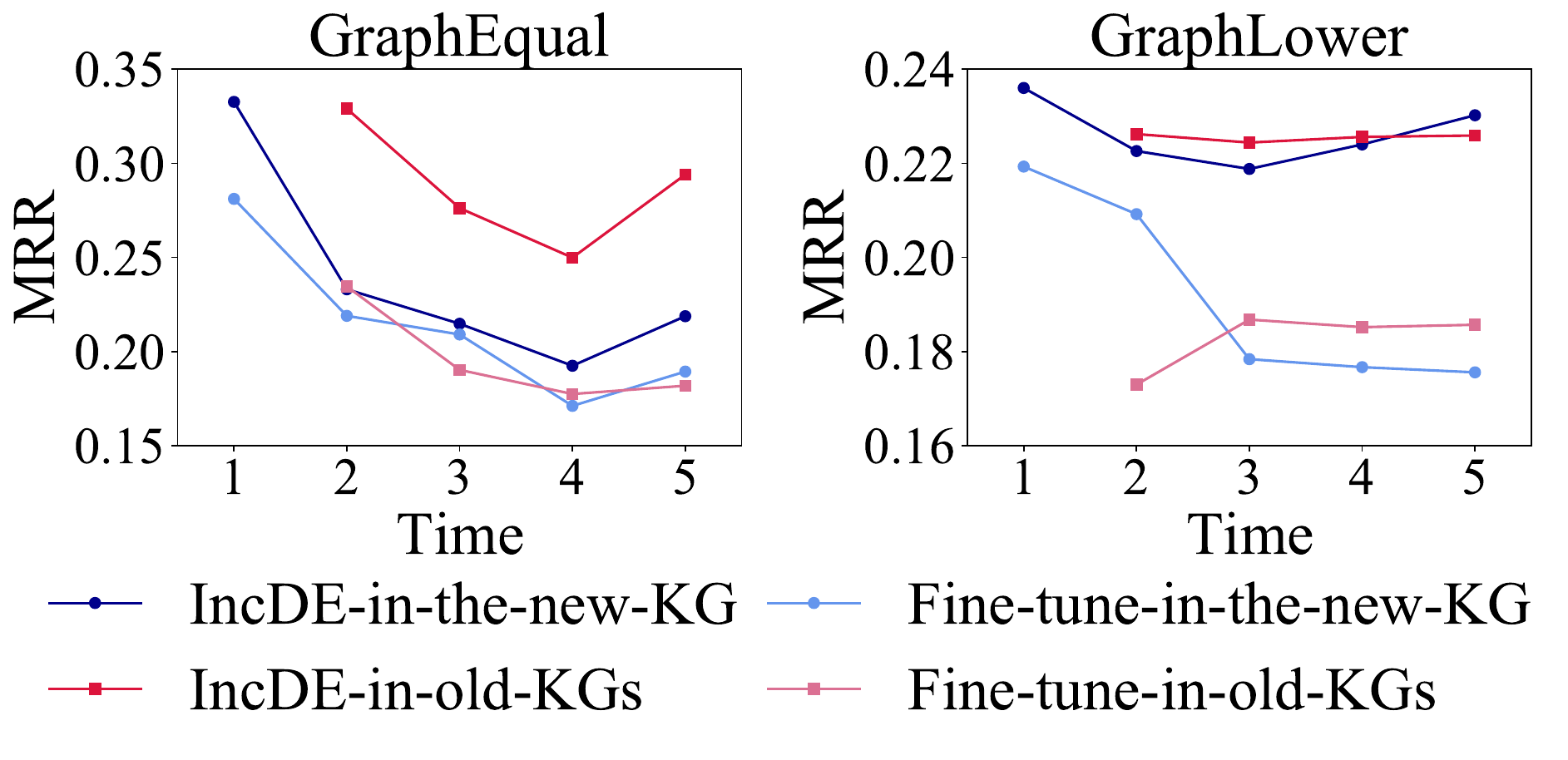}
\caption{Effectiveness of learning emerging knowledge and memorizing old knowledge.}
\label{p7}
\end{figure}

\begin{figure}[t]
\includegraphics[width=0.47\textwidth]{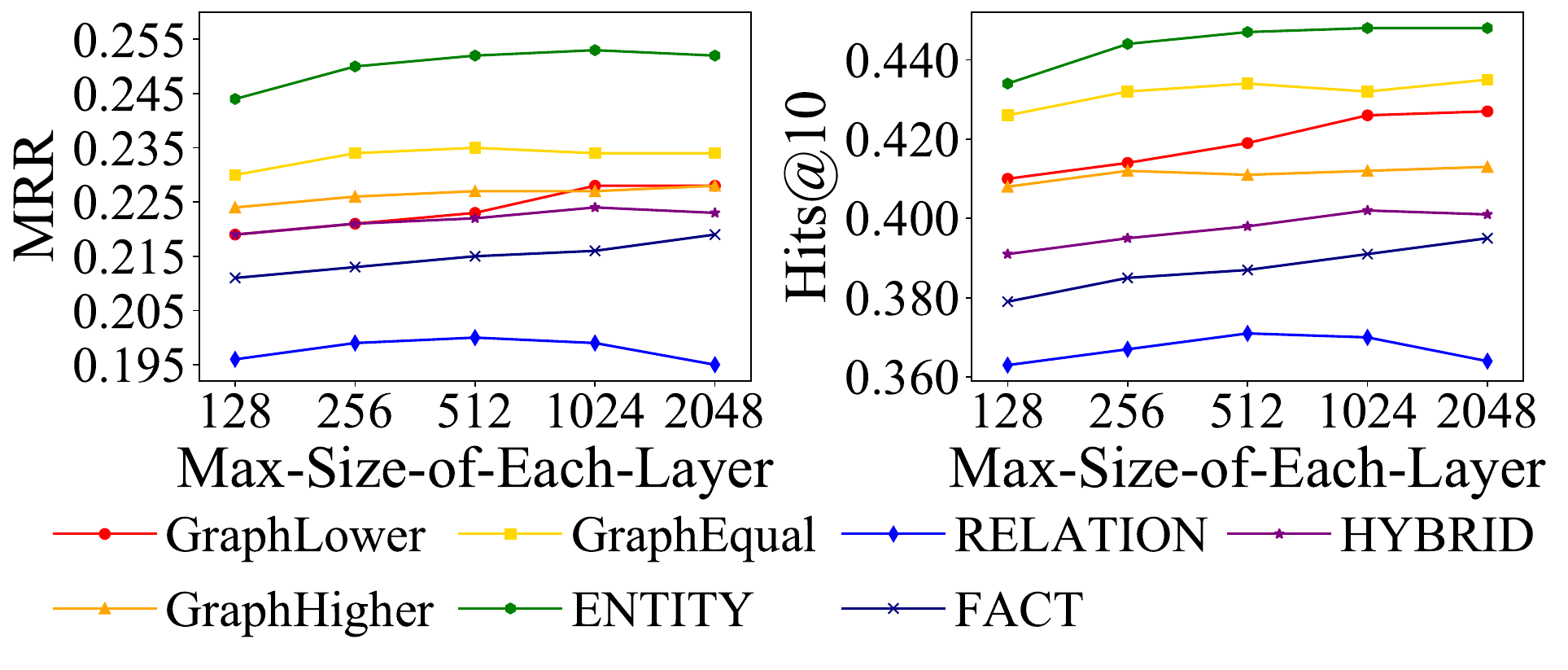}
\caption{Results of MRR and Hits@10 with different max sizes of layers in all datasets.}
\label{figure4}
\end{figure}

\subsubsection{Effect of Learning and Memorizing}
In order to verify that IncDE can learn emerging knowledge well and remember old knowledge efficiently, we study the effect of IncDE and Fine-tune each time on the new KG and old KGs, respectively, as shown in Figure~\ref{p7}. 
To assess the performance on old KGs, we calculated the mean value of the MRR across all past time steps. 
Firstly, we observe that IncDE outperforms Fine-tune on the new KG, with a higher MRR ranging from 0.5\% to 5.5\%. 
This indicates that IncDE is capable of effectively learning emerging knowledge. 
Secondly, IncDE has 3.8\%-11.2\% higher than Fine-tune on old KGs in MRR. 
These findings demonstrate that IncDE mitigates the issue of catastrophic forgetting and achieves more efficient retention of old knowledge.

\subsubsection{Effect of Maximum Layer Sizes}
To investigate the effect of the max size of each layer $M$ in incremental distillation on model performance, we study the performances of IncDE models at the last time with different $M$, as shown in Figure~\ref{figure4}. 
First, we find that the model performance on all datasets rises with $M$ in the range of [128, 1024]. 
This indicates that, in general, the higher $M$, the more influential the incremental distillation becomes. 
Second, we observe a significant performance drop on some datasets when $M$ reaches 2048. 
It implies that too large an $M$ could lead to too few layers and limit the performance of incremental distillation. 
Empirically, $M$=1024 is the best size in most datasets. 
This further proves that it is necessary to limit the number of triples learned in each layer.

\begin{table}[t]
\setlength\tabcolsep{0.9pt}
\centering
\small
\begin{tabular}{cc}
\hline
Query & \textit{(Arizona State University, major\_field\_of\_study, ?)} \\
\hline
\hline
 Methods & Top 3 Candidates \\
\hline
 EWC & Medicine, Electrical engineering, \textbf{Computer Science} \\
 PNN & Medicine, Electrical engineering, \textbf{Computer Science} \\
 LKGE & English Literature, \textbf{Computer Science}, Political Science \\
\hline
 IncDE & \textbf{Computer Science}, University of Tehran, Medicine \\
 w/o HO & \textbf{Computer Science}, Medicine, University of Tehran \\
 w/o ID & Political Science, English Literature, Theatre \\
 w/o TS & \textbf{Computer Science}, Medicine, University of Tehran \\
\hline
\end{tabular}
\caption{Results of the case study. We use the model generated at time 5 and randomly select a query appearing in ENTITY at time 1 for prediction. The italic one is the query, and the bold ones are true prediction results.}
\label{t6}
\end{table}

\subsubsection{Case Study}
To further explore the capacity of IncDE to preserve old knowledge, we conduct a comprehensive case study as shown in Table \ref{t6}. 
In the case of predicting the \textit{major field of study} of \textit{Arizona State University}, IncDE ranks the correct answer \textit{Computer Science} in the first position, outperforming other strong baselines such as EWC, PNN, and LKGE, which rank it second or third. 
It indicates that although other methods forget knowledge in the past time to some degree, IncDE can remember old knowledge at each time accurately. 
Moreover, when incremental distillation (ID) is removed, IncDE fails to predict the correct answer within the top three positions. 
This demonstrates that the performance of IncDE significantly declines when predicting old knowledge without the incremental distillation. 
Conversely, after removing hierarchical ordering (HO) and the two-stage training strategy (TS), IncDE still accurately predicts the correct answer in the first position. 
This observation strongly supports the fact that the incremental distillation provides IncDE with a crucial advantage over alternative strong baselines in preserving the old knowledge.

\subsection{Discussion}
\subsubsection{Novelty of IncDE}
The novelty of IncDE can be summarized by the following two aspects. 
(1) \textit{Efficient knowledge-preserving distillation.} 
Although IncDE utilizes distillation methods, it is different from previous KGE distillation methods~\cite{wang2021mulde, zhu2022dualde, liu2023iterde}. 
For one thing, compared to other KGE distillation methods that mainly distill final distribution, incremental distillation (ID) distills the intermediate hidden states. 
Such a manner skillfully preserves essential features of old knowledge, making it adaptable to various downstream tasks. 
For another thing, only ID transfers knowledge from the model itself, thus mitigating error propagation compared to transferring knowledge from other models. 
(2) \textit{Explicit graph-aware mechanism.}
Compared to other CKGE baselines, IncDE stands out by incorporating the graph structure into continual learning. 
This explicit graph-aware mechanism allows IncDE to leverage the inherent semantics encoded within the graph, enabling it to intelligently determine the optimal learning order and effectively balance the preservation of old knowledge. 

\subsubsection{Three Components in IncDE}
The three components of IncDE, hierarchical ordering (HO), incremental distillation (ID), and two-stage training (TS) are inherently dependent on each other and necessary to be used together. 
We explain it in the following two aspects. 
(1) \textit{Designing Principle.}
The fundamental motivation of IncDE lies in effectively learning emerging knowledge while simultaneously preserving old knowledge. 
This objective is accomplished by all three components: HO, ID, and TS. 
On the one hand, HO plays a role in dividing new triples into layers, optimizing the process of learning emerging knowledge. 
On the other hand, ID and TS try to distill and preserve the representation of entities, ensuring the effective preservation of old knowledge. 
(2) \textit{Inter Dependence.}
The three components are intrinsically interrelated and should be employed together. 
For one thing, HO plays a vital role in generating a partition of new triples, which are subsequently fed into ID. 
For another thing, by employing TS, ID prevents old entities from being disrupted in the early training stages. 

\subsubsection{Significance of Incremental Distillation}
Even though the three proposed components of IncDE: incremental distillation (ID), hierarchical ordering (HO), and two-stage training (TS) are all effective for the CKGE task, ID serves as the central module among them. 
Theoretically, the primary challenge in the continual learning task is catastrophic forgetting that occurs when learning step by step, which is also suitable for the CKGE task. 
To tackle this challenge, ID introduces the explicit graph structure to distill entity representations, effectively preserving old knowledge layer by layer during the whole training time. 
However, HO focuses on learning new knowledge well, and TS can only alleviate catastrophic forgetting in the early stages of training. 
Therefore, ID plays the most important role among all components in the CKGE task. 
In experiments, we observe that ID exhibits significant improvements (4.1\% in MRR on average) compared to HO (0.9\% in MRR on average) and TS (0.5\% in MRR on average) from Table~\ref{t4} and Table~\ref{t5}. 
Such results further verify ID as the pivotal component compared with HO and TS. 
The three components interact with each other and work together to complete the CKGE task. 

% \subsubsection{Scalability of IncDE}

\section{Conclusion}
This paper proposes a novel continual knowledge graph embedding method, IncDE, which incorporates the graph structure of KGs in learning emerging knowledge and remembering old knowledge. 
Firstly, we perform hierarchical ordering for the triples in the new knowledge graph to get an optimal learning sequence. 
Secondly, we propose incremental distillation to preserve old knowledge when training the new triples layer by layer. 
Moreover, We optimize the training process with a two-stage training strategy. 
In the future, we will consider how to handle the situation where old knowledge is deleted as knowledge graphs evolve. 
Also, it is imperative to address the integration of cross-domain and heterogeneous data into expanding knowledge graphs.

\section{Acknowledgments}
We thank the anonymous reviewers for their insightful comments. 
This work was supported by National Science Foundation of China (Grant Nos.62376057) and the Start-up Research Fund of Southeast University (RF1028623234). 
All opinions are of the authors and do not reflect the view of sponsors.

\bibliography{aaai24}

\end{document}